\documentclass[letterpaper, 10 pt, conference]{ieeeconf}  % Comment this line out if you need a4paper

\IEEEoverridecommandlockouts                              % This command is only needed if 
\usepackage{mathptmx} % assumes new font selection scheme installed
\usepackage{amsmath} % assumes amsmath package installed
\usepackage{amssymb}  % assumes amsmath package installed
\usepackage{graphicx}
\usepackage{multirow}
\usepackage{booktabs}
\usepackage{wrapfig}
\usepackage{mathtools}
\usepackage{subcaption}
\usepackage{url}
\usepackage{algorithm}
\usepackage{algpseudocode}
\usepackage{hyperref}

\algblock{Input}{EndInput}
\algnotext{EndInput}
\algblock{Output}{EndOutput}
\algnotext{EndOutput}
\newcommand{\Desc}[2]{\State \makebox[3em][l]{#1}#2}

\title{\LARGE \bf
\method: Multi-agent Task Planning using Classical Planners and Large Language Models
}

\author{David Bai$^{*1}$, Ishika Singh$^{*1}$, David Traum$^{1}$, Jesse Thomason$^{1}$% <-this % stops a space
\thanks{*Equal Contribution}% <-this % stops a space
\thanks{$^{1}$Computer Science Department, University of Southern California {\tt\small \{dmbai, ishikasi, traum, jessetho\}@usc.edu}}%
}

\newcommand{\method}[0]{\textsc{TwoStep}}
\newcommand{\sa}[0]{\textsc{SA PDDL}}
\newcommand{\ma}[0]{\textsc{MA PDDL}}
\newcommand{\human}[0]{\textsc{TwoStep}-Human}
\newcommand{\barman}[0]{\texttt{BARMAN}}
\newcommand{\blocks}[0]{\texttt{BLOCKSWORLD}}
\newcommand{\termes}[0]{\texttt{TERMES}}
\newcommand{\tyre}[0]{\texttt{TYREWORLD}}
\newcommand{\gripper}[0]{\texttt{GRIPPERS}}
\newcommand{\thor}[0]{\texttt{AI2THOR}}

\newcommand{\helper}[0]{\texttt{helper}}
\newcommand{\main}[0]{\texttt{main}}

\usepackage{xcolor}
\usepackage{colortbl}
\newcommand{\good}[1]{\textcolor{blue}{#1}}
\newcommand{\bad}[1]{\textcolor{red}{#1}}
\definecolor{Gray}{gray}{0.9}
\newcolumntype{o}{>{\columncolor{Gray}}l}
\newcolumntype{k}{>{\columncolor{Gray}}r}

\begin{document}
\maketitle
\thispagestyle{empty}
\pagestyle{empty}

%%%%%%%%%%%%%%%%%%%%%%%%%%%%%%%%%%%%%%%%%%%%%%%%%%%%%%%%%%%%%%%%%%%%%%%%%%%%%%%%
\begin{abstract}
Classical planning formulations like the Planning Domain Definition Language (PDDL) admit action sequences guaranteed to achieve a goal state given an initial state if any are possible. 
However, reasoning problems defined in PDDL do not capture temporal aspects of action taking, such as concurrent actions between two agents when there are no conflicting conditions, without significant modification and definition to existing PDDL domains.
A human expert aware of such constraints can decompose a goal into subgoals, each reachable through single agent planning, to take advantage of simultaneous actions.
In contrast to classical planning, large language models (LLMs) directly used  for inferring plan steps rarely guarantee execution success, but are capable of leveraging commonsense reasoning to assemble action sequences.
We combine the strengths of both classical planning and LLMs by approximating human intuitions for multi-agent planning goal decomposition.
We demonstrate that LLM-based goal decomposition leads to faster planning times than solving multi-agent PDDL problems directly while simultaneously achieving fewer plan execution steps than a single agent plan alone, as well as most multiagent plans, while guaranteeing execution success.
Additionally, we find that LLM-based approximations of subgoals result in similar multi-agent execution lengths to those specified by human experts. 
Website and resources at \href{https://glamor-usc.github.io/twostep}{glamor-usc.github.io/twostep}
\end{abstract}

%%%%%%%%%%%%%%%%%%%%%%%%%%%%%%%%%%%%%%%%%%%%%%%%%%%%%%%%%%%%%%%%%%%%%%%%%%%%%%%%
\section{Introduction}

Symbolic planning problems specified in PDDL tend to explore single agent plans, and changing these domain specifications to enable planning with multiple agents requires a human expert.
Unfortunately, multi-agent planning introduces an exponential growth in the search space over possible plans, since each agent can take an action at each timestep.
In this paper, we explore whether the commonsense reasoning abilities in LLMs can take advantage of the lexical semantics encoded in the variable, action, and other ontological names of an expert-written planning domain to predict subgoals for individual agents that still achieve a given global goal.
We present our study in both purely symbolic execution domains as well as embodied simulation domains. 

% TEACh~\cite{teach} proposes an embodied multi-agent benchmark with a commander and a follower agent. 
% The commander can query for oracle information which it can communicate through dialogue to the follower, which in turn takes embodied actions to complete the task. 
% We propose a collaborative embodied AI framework, where all agents share and operate in the same environment.

\textbf{Classical planning.}
Classical or automated task planning algorithms have been widely applied in autonomous spacecrafts, military logistics, manufacturing, games, and robotics. 
The automated STRIPS planner, for example, operated the Shakey robot~\cite{strips} in 1970.
Classical planning algorithms work with finite, deterministic, and full state information, and guarantee finding a plan when a path from the initial to goal states is possible. 
Some other frameworks have also shown to be useful for robot planning that use hierarchical task structures~\cite{prodigy, htn}.
Planning domain description language (PDDL) and answer set programming (ASP) are popular specification formats for planning domains~\cite{pddl, pddl1, asp, asp1}. 
% Task and Motion Planning is a hierarchical planner that provides both discrete state classical planning and continuous state motion planning~\cite{tamp, tamp1}.

\textbf{Planning  with LLMs.}
Several works have shown that LLMs can guide agent actions by acting as planners~\cite{zeroshot, saycan}, but such stochastic, generative approaches lose the success guarantees of classical planners. 
Another line of work shows that LLMs plan better with API-based programmatic plan generation~\cite{progprompt, cap, progprompt-1}, which introduces some symbolic structure and constraints but still does not ensure success. 
Program synthesis for planning has been previously proposed in LEAPS~\cite{trivedi2021leaps}, which generates programs using a learned latent space. 
PDDL can also be used as a translation target for natural language, with LLMs used to generate either a PDDL plan or goal~\cite{pddlplanning,llmp}.
Generating a PDDL goal eliminates the need to track or predict world state after each action, which is required when using LLMs for open- or closed-loop planning, respectively..
However, using a classical planner necessitates specification of all domain information.
LLMs have also been used for multi-agent planning through code generation~\cite{SMARTLLMSM} and task allocation through dialogue and path planning~\cite{roco}.
In this paper, we use LLMs to infer commonsense ways to decompose a goal, but leave formal planning to a PDDL system to ensure plan correctness, using each tool for what it is better at.

\section{Background}

We formally define the planning problem for single and multi-agent settings and provide an overview of PDDL.

\subsection{Classical planning problem}
Planning problem $\mathrm{P}$ is formally defined as a tuple,
\begin{align*}
    \langle\mathcal{S}, s^{i}, \mathcal{S}^g, \mathcal{A}, \mathcal{O}, \mathcal{P}, \mathcal{T}\rangle,
\end{align*}
where: $\mathcal{S}$ is a finite and discrete set of environment states; $s^i \in \mathcal{S}$ is the initial state; $\mathcal{S}^g \subset \mathcal{S}$ is the set of goal states that satisfy a goal conditions $g$; and $\mathcal{A}, \mathcal{O}, \mathcal{P}$ are sets of symbolic actions, objects, and environment predicates, respectively. 
A state is defined as a list of environment predicates taking object and agent arguments that hold true. 
The underlying transition function $\mathcal{T}$ defines how the state conditions change when an action $a \in \mathcal{A}$ is executed in the environment.
A solution to $\mathrm{P}$ is a symbolic plan $\pi$ given as a sequence of actions  $a_t|_{t=1-T}$, such that preconditions of each $a_t$ holds in the state $s_{t-1}$, where $s_0 = s^t$ and $s_N \in \mathcal{S}^g$.

\begin{figure}[h!]
    \centering
    \begin{subfigure}[t]{0.24\textwidth}
        \centering
        \includegraphics[width=0.9\textwidth]{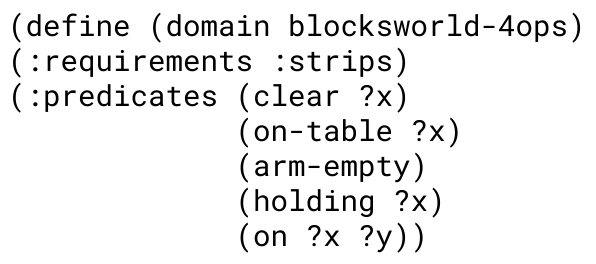}
        \caption{PDDL domain description}
        \label{pddl-1}
    \end{subfigure}%
    ~ 
    \begin{subfigure}[t]{0.25\textwidth}
        \centering
        \includegraphics[width=0.9\textwidth]{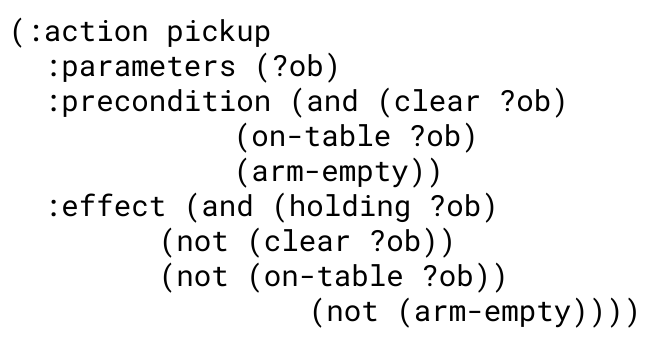}
        \caption{PDDL action operator}
        \label{pddl-2}
    \end{subfigure}
    ~ 
    \begin{subfigure}[t]{0.26\textwidth}
        \centering
        \includegraphics[width=0.9\textwidth]{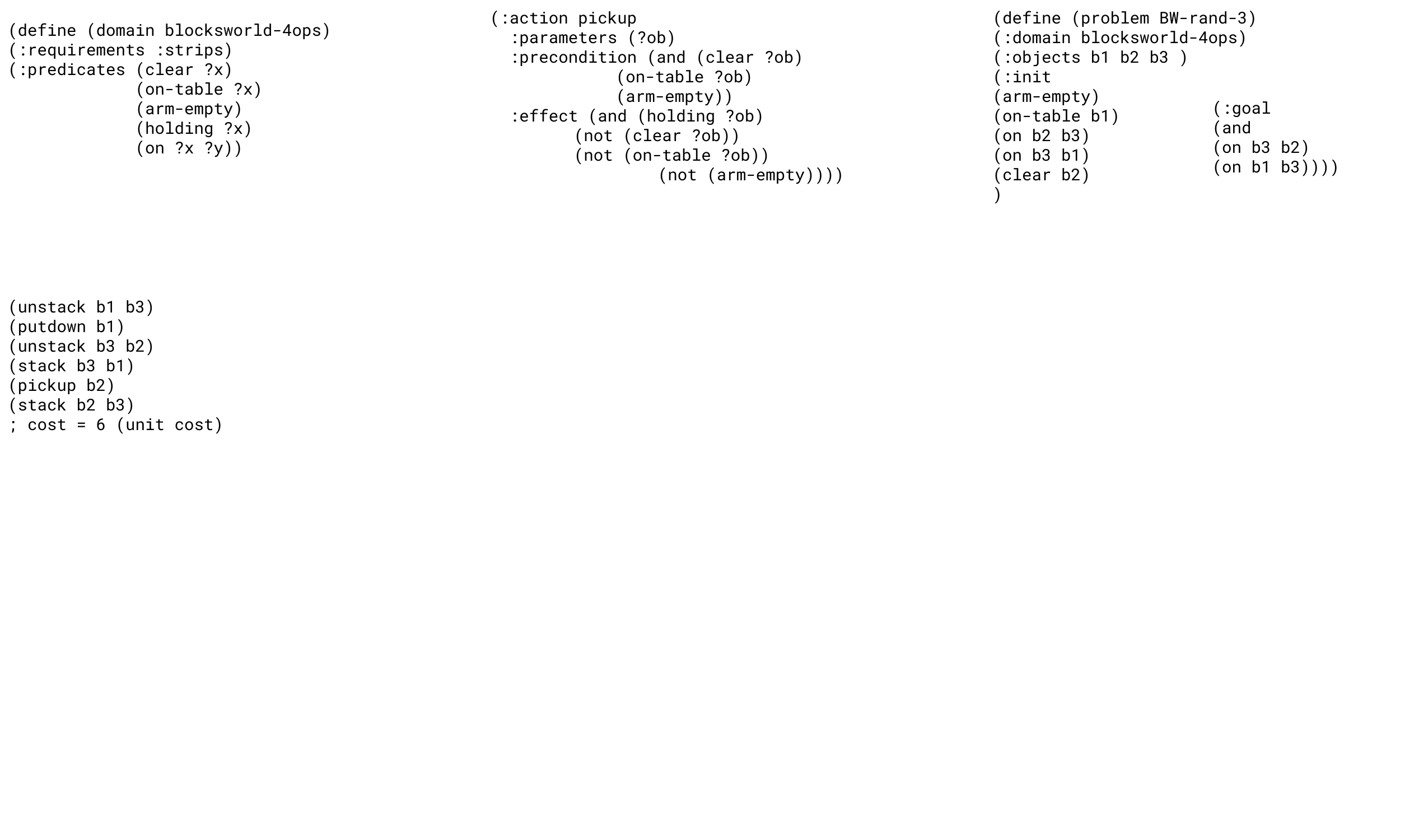}
        \caption{PDDL problem}
        \label{pddl-3}
    \end{subfigure}
    ~ 
    \begin{subfigure}[t]{0.17\textwidth}
        \centering
        \includegraphics[width=0.9\textwidth]{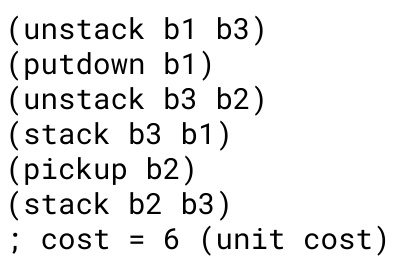}
        \caption{PDDL plan}
        \label{pddl-4}
    \end{subfigure}
    \caption{PDDL overview. 
    PDDL planning requires a domain definition, with state predicates (a) and action operators (b), and an initial and goal condition specifications (c). 
    A solver can then use these definitions to generate a plan (d).}
\end{figure}

\subsection{Planning Domain Definition Language (PDDL)}
Planning Domain Definition Language (PDDL) serves as a standardized encoding language to represent planning problems.
In this paper, we consider deterministic, fully-observable planning problems. 
The planning problem in PDDL is represented by two files: domain and problem. The domain file defines the environment in terms of the object types and predicates (Figure~\ref{pddl-1}), followed by action and their parameters, required preconditions, and postcondition effects (Figure~\ref{pddl-2}). 
Some conditions relate to the agent \texttt{\mbox{(holding ?x)}} and others to the environment \texttt{\mbox{(on-table ?x)}}. 
The problem file defines an instance of the domain with a list of objects, initial state and goal conditions (Figure~\ref{pddl-3}). 
We use a PDDL solver~\cite{fd} that takes in a domain and problem file and searches for action sequences leading from the initial state to a goal state, heuristically returning the shortest such plan found, if any (Figure~\ref{pddl-4}).

\subsection{Multi-agent planning problem}
\label{sec: RW-MAplanning}
In multi-agent settings, actions are executed by the multiple agents in parallel. 
In this case, agents share the same environment, so the environment-specific state conditions are shared between the agents.
However, agent-specific state conditions are unique to each agent. 
Agents must consider environment conditions, their own conditions, and other agent's conditions to execute a coordinated action.
A multi-agent plan consists of actions for each agent at every time step $t$, $a_t^r|_{t=1-T} ^{r=1-|R|}$, for $r\in R$ agents.
Multiagent PDDL planners may or may not support parallel execution aware planning, such that multiple agents can take an action in a single timestep.
The PDDL implementations and solvers that have this feature required involved modification to the PDDL domains and solvers through methods such as temporal constraints and definitions per action\cite{Eyerich_Mattmüller_Röger_2009}, as well as new types of objects modeling different agent states, per agent \cite{Furelos-Blanco_Jonsson_2019}.

\newcommand{\sggen}[0]{subgoal generator}
\newcommand{\sgtrans}[0]{subgoal translator}
\newcommand{\engsg}[0]{English subgoal}
\newcommand{\pddlsg}[0]{PDDL subgoal}

\section{Multi-agent Planning Method: \method}

\begin{figure*}[t]
  \centering
  \includegraphics[width=\textwidth]{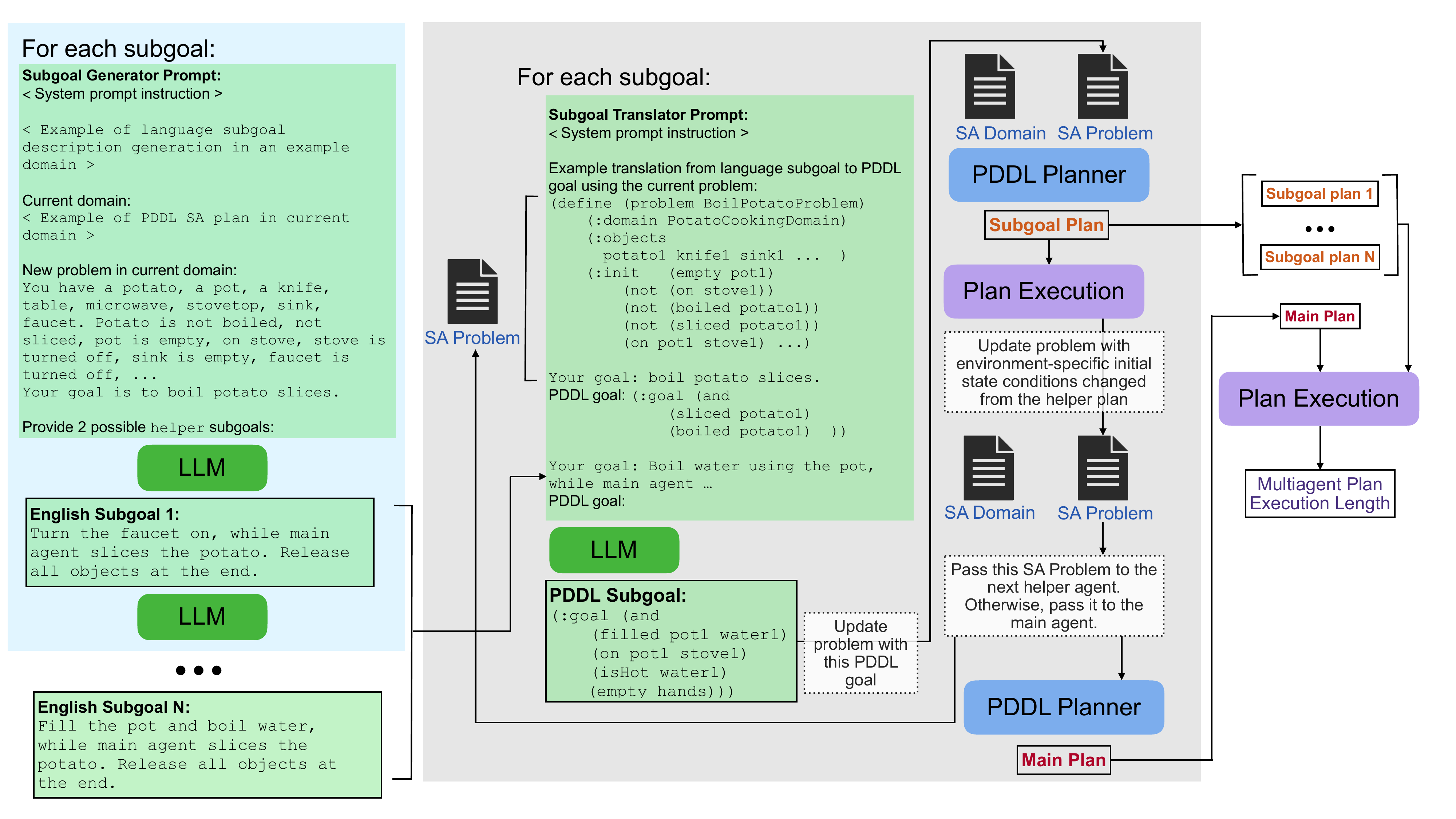}
  \caption{
  Planning pipeline for \method. 
  \method\ has $N$ \helper\ agents and a \main\ agent. Each \helper\ agent tries to extract and complete a partially independent subgoal that reduces the steps for \main\ agent, while \main\ agent completes the remaining task after all helpers have completed subgoals.
  The plans generated for all agents are then executed to find the execution length.}
  \label{fig:pipeline}
\end{figure*}

We propose decomposing an $N$-agent planning problem into $N$ single-agent planning problems by leveraging the human-like commonsense reasoning capabilities of LLMs. 
Specifically, we consider a multiagent scenario with $N-1$ \helper\ agents and one \main\ agent.
For a given problem $\mathrm{P}$, containing object definitions, initial state $i$ and goal conditions $g$, each \helper\ agent $h$ generates a plan $\pi_h = \prod(i, g_h)$ to reach a subgoal state $g_h$ from the initial state $i$ using a planner $\prod$. The resulting state $i_{h+1} = E(i, \prod(i, g_h))$, where $E$ is \textit{Plan Execution} returning the state reached after starting from $i$ and executing steps $\prod(i, g_h)$, serves as the starting point for the next \helper\ agent.
This iteration continues until the \main\ agent executes $\pi_m = \prod(i_{N-1}, g)$ to achieve the final goal $g$. 
Each \helper\ agent’s subgoal $g_h$ is generated through two modules: \sggen\ for English subgoal generation and \sgtrans\ for translating it into a \pddlsg. We hypothesize that LLMs can infer \helper\ subgoals that enable parallel execution alongside the \main\ agent while assuming all agents will eventually achieve their respective goals.
We introduce \method\ (Figure~\ref{fig:pipeline}) to evaluate this hypothesis.

\paragraph{English subgoal generation for \helper\ agents}
Given an in-context \engsg\ example and a new problem $P^{d}$ in domain $d$, the LLM iteratively generates \engsg s for $N-1$ \helper\ agents.
We construct an in-context example problem $P_{eg}^{d'}$ in a fixed training domain $d'$ with an initial state $i_{eg}^{d'}$, goal conditions $g_{eg}^{d'}$, and a corresponding single-agent PDDL plan $\pi_{eg}^{d'} (i_{eg}^{d'}, g_{eg}^{d'})$. We then introduce a new problem $P^{d'}$ in the same domain with no associated plan. Next, we present two potential \engsg s for \helper\ agents for the in-context example, ensuring intermediate progress toward $g^{d'}$ while promoting independent execution for concurrent subgoal completion. These subgoals in the prompt also encourage that objects are released after use to prevent blocking the \main\ agent (e.g., avoiding one \helper\ holding a pot needed by the \main\ agent in Figure~\ref{fig:pipeline} (left)).

To solve the current problem $P^{d}$, we further append the LLM prompt to include the current domain’s ($d$) description, an example problem $P_{eg}^{d}$ with its single-agent plan $\pi_{eg}^{d} (i_{eg}^{d}, g_{eg}^{d})$, and the new problem $P^{d}$. The LLM iteratively generates \engsg s until $N-1$ subgoals are obtained. The prompt also enables the LLM to determine that no more \helper\ agents are needed by outputting `None'. We provide the full prompt on the project website.

\paragraph{Translation from English description of subgoals to PDDL goal specifications}

To translate \engsg s into \pddlsg s, we provide the LLM with the problem $P^d$ and initial state $i^d$ in PDDL. Next, we append an example mapping of an English goal description to its PDDL equivalent $g^d$ before prompting the LLM to translate the generated \engsg s into \pddlsg s.
By using in-context examples from the current problem, we ensure that the translated subgoals $g^d_h|_1^{N-1}$ are valid within the problem instance and executable by a planner. If a subgoal fails to generate a plan within a given time constraint, we discard it from further processing. 

\paragraph{Editing initial state for \main\ agent}

To account for changes in the environment state caused by the $N-1$ \helper\ agents' plans, $\pi^d_{h}$, we iteratively update the initial state. We compute the subgoal states $g^d_h$ by applying the effects of the \helper\ agents' actions. Starting from the current problem’s initial state $i^d$, we update the environment conditions after each \helper\ agent’s plan $\pi^d_h$ is executed.
These state updates consider not only the final \pddlsg\ achieved by each \helper\ agent but also any intermediate state changes occurring during the execution of plans, ensuring smooth state transitions between agents and aligning interfacing states $g^d_h$ and $i^d_{h+1}$ to maintain consistency. Importantly, we do not modify agent-specific conditions in $i^d$ so that each \helper\ agents, as well as the \main\ agent, can start from their own respective initial states. This choice allows agents to operate within a partially solved environment while still fulfilling the remaining conditions required to achieve the original goal $g^d$.

\paragraph{Multi-agent plan execution}

Each agent's plan begins from the previous agent's final environment state. \method\ encourages the LLM to generate $N-1$ \helper\ subgoals that can run in parallel with the \main\ agent. However, agents may need to wait for others to complete certain actions before proceeding.
To ensure efficient execution, we maintain a shared environment state initialized from $i^d$ while tracking individual agent states separately. Our results show that partial parallel execution of subgoals effectively reduces overall plan length.

\section{Experiment setup}

Our principal experiments measure whether \method\ can lead to more efficient multi-agent tasks than a solution to single-agent and multi-agent PDDL problems in terms of agent actions, while simultaneously requiring less planning time than needed to solve a multi-agent PDDL problem formulation. We demonstrate \method\ with up to 4 agents and find that the LLM, as mentioned in Section III.a, often decides not to generate more subgoals than 3 (which when including the main goal, represents 4 agents), even when given the option to. 
Overall, we find that \method\ achieves these results when both the planning domain considers agent state in action preconditions, and the planning problem to solve contains two or more partially independent subgoals rather than requiring strict action sequencing.

\subsection{Domains}

We evaluate \method\ on 5 symbolic domains utilized in prior work that leveraged LLMs for planning~\cite{llmp} and in one simulation environment~\cite{ai2thor}.

\subsubsection{Symbolic Domains}
Each symbolic environment defines 20 tasks that vary in number of the environment objects and optimal plan length.
These domains are frequently used in classical planning competitions. 
1) \barman: A bartender with 2 hands has been assigned the job of preparing cocktails for a customer's order, using the specified ingredient and appropriate tools.\footnote{
% , and striving to accomplish this with the fewest possible actions.
For all experiments, we slightly modify \barman\ by increasing the number of shakers from one to three, enabling the possibility of parallelization.
}
2) \blocks: A collection of block piles arranged on a table, and a robot is tasked to reorganize them into a specified configuration while adhering to the simple physics principles. 
3) \termes: The objective of a robot is to construct complex structures by transporting and positioning blocks, as well as using them as a means to ascend and move adjacent blocks. 
4) \tyre: The robot is assigned with changing flat tires, which involves tasks such as removing flat tires, inflating the intact tires, tightening nuts, and returning tools to the boot, all in the correct sequence.\footnote{
\tyre\ has a quirk compared to other domains in that its preconditions do not consider agent state.
For example, the agent need not be holding a tool first in order to apply the tool to a tire.
Tools in the environment are ``available'' to the agent without explicitly taking posession of them first, resulting in less constrained multiagent coordination.}
5) \gripper: A set of robots with 2 grippers each are given a task to move objects among different rooms.

\subsubsection{Embodied Domain} 
We use \thor\ simulator for multi-agent task execution.
We conduct planning using \texttt{AlfWorld}, a text-based abstraction of \thor\ that includes only high-level actions without low-level execution details.
\texttt{AlfWorld} provides the domain and tasks encoded in PDDL. 
% We extend the support to multi-agent PDDL planning, by initializing multiple agents in the PDDL initial environment state as well as the simulator. 

The multi-agent simulation, \thor, is implemented asynchronously, i.e., each agent gets to take an action at every time step, however, due to the underlying \thor\ API, only one agent can be issued a command each simulator timestep.
We build a parallel plan execution framework that allows action queuing if an agent is occupied with a previous action, such as ‘go to microwave 1’ which involves multiple navigation steps. 
The oracle multi-agent navigation pipeline uses object location and position and other agent's position to navigate, abstracting away low-level detail similar to \texttt{AlfWorld}.
Deadlocks during agent navigation are resolved with a simple random backoff strategy where one agent is randomly selected to move to a random reachable position first before re-attempting navigation to its next goal.
Reachable positions, where an agent can navigate to, at any time change when another agent moves in the environment as it occupies space, akin, to any other object.
% The simulation supports path replanning around another agent while navigating, however, there are scenarios where the agents are deadlocked together if there's no way around the collision. 
% At this point, we defer a randomly chosen agent to a randomly chosen location to relieve from the deadlock. 
% Thereafter, the agents continue to execute their queued actions.

\begin{algorithm}[h]
\caption{Parallel Execution Length Computation Algorithm for symbolic domains}
\label{alg:exec}
\begin{algorithmic}[1]
\Input
  \Desc{$\pi$}{ array of plans for all agents}
  \Desc{$indices$}{ current step in the plan per agent}
  \Desc{$i$}{ specific agent to execute}
  \Desc{$n$}{ number of agents}
  \EndInput
\Output
  \Desc{$length$}{min. execution length for given indices}
  \EndOutput
\Function{Solve}{$\pi, indices,i=None$}
    \If {$\forall j, indices[j] = \lvert \pi[j] \rvert$}
        \State \Return $0$\
    \EndIf
    \If {$i$ is not None} 
        \State $V_i \gets execute \ \pi[i][indices[i]]$
        \If {$V_i$}
        \State $\triangleright$ Advance agent index, get cost of next step
        \State $new\_indices[i]\gets indices[i]+1$
        \State \Return $1 + $
        \Call{Solve}{$new\_indices, plans$}
    \Else
        \State \Return $\infty$
    \EndIf
    \EndIf
    \If{$i$ is None}\
        \State $\triangleright$ Get costs for all single agent executions
        \State $\text{SA\_res}_j \gets [\Call{Solve}{\text{indices}, \text{plans}, j}]_{j=1}^n$
        \State $V_{\text{all}} \gets execute\  \pi_1[\text{indices}[1]], \ldots, \pi_n[\text{indices}[n]]$
        \If{$V_{\text{all}}$}
            \State $\triangleright$ If possible, execute all agents in parallel
            \State $new\_indices \gets increment\_all\_indices$
            \State $\text{PA\_res}\gets \Call{Solve}{new\_indices,plans}$
        \Else
            \State $\text{PA\_res}\gets \infty$
        \EndIf
        \State \Return $1 + \min($\text{PA\_res}, \text{SA\_res})
    \EndIf
\EndFunction
\end{algorithmic}
\end{algorithm}

\subsection{Approaches Considered}
We compare \method\ against single agent and multi-agent PDDL planning for all the above domains. 
We also define 4 long-horizon tasks and their corresponding PDDL problem files for our evaluations in the embodied domain (Table~\ref{tab:alfworld}).
Some objects may also be inside another containers like drawers and cabinets as per the PDDL initial state specification, making the resultant plans longer. 
We will open-source our defined domain and problem files.
We consider the following approaches for the above domains,

\begin{figure*}[ht]
  \centering
  \includegraphics[width=\textwidth]{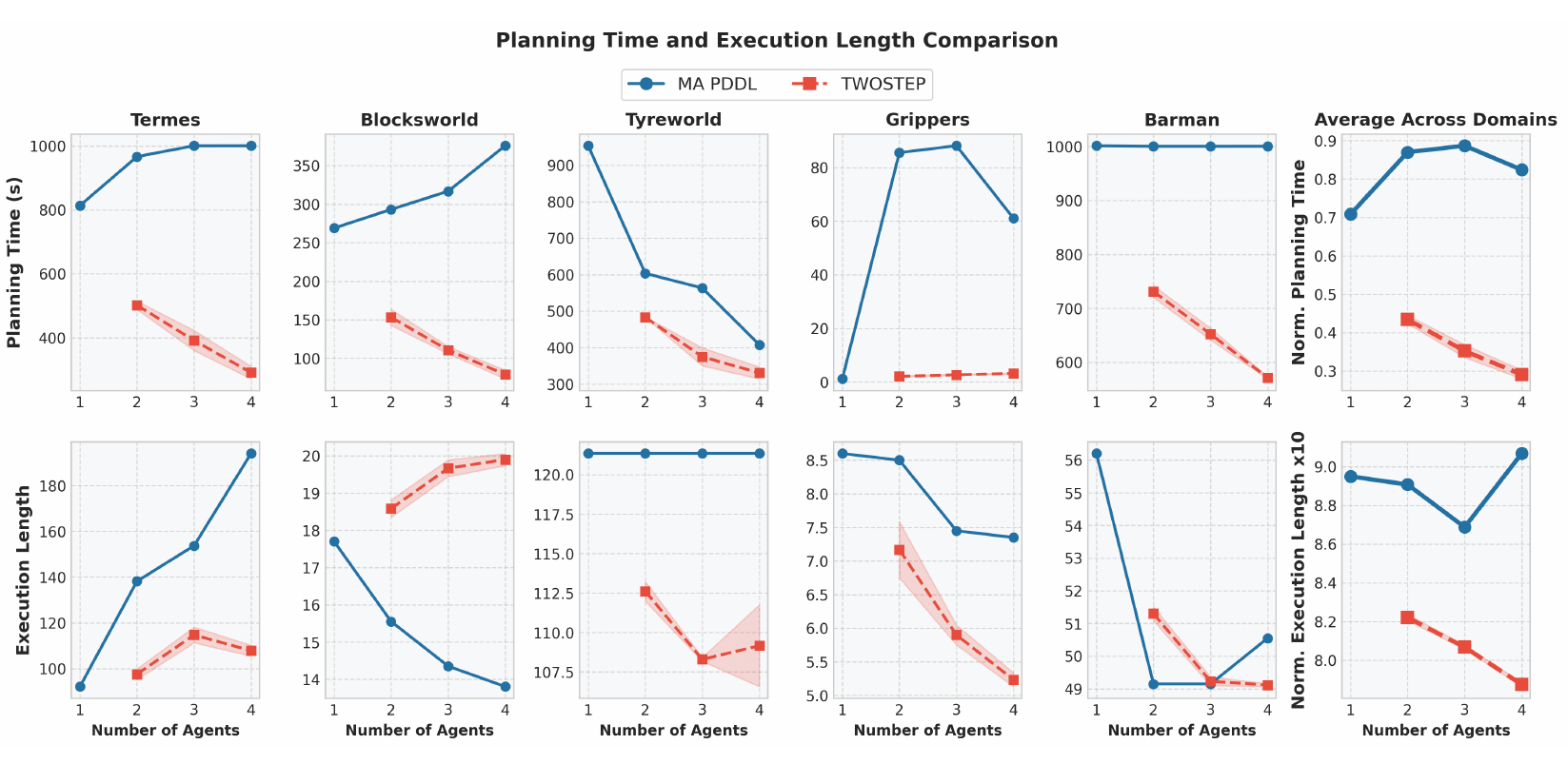}
  \caption{
    Results comparing vanilla planning from \sa\ ($N=1$) and \ma\ ($N\geq1$) to \method. The shaded background is the standard deviation across 3 runs for \method, and the rightmost graphs represent the average of the normalized per domain results across all domains. All averages are across all 20 tasks, with the exception of \ma\ in the \termes\ domain, where planning times without producing a plan out for 5/20 tasks with 4 agents. 
    % On the other hand, \method\ allows us to execute all tasks for 2 through any $N$ agents, with 1 agent execution being equivalent to \sa.
    On average, \method\ reduces planning time and improves parallelization over vanilla multi-agent PDDL planning as the number of agents increases.
    }
  \label{fig:results}
\end{figure*} 

1) \sa: vanilla PDDL planning for a single agent (SA). This baseline will compare the proposed \method\ with a single agent completing the task using a single agent plan generated by the PDDL planner where domain files have been adjusted to enable agent-specific actions.

2) \ma: PDDL planning in the above domains for two or more agents. This baseline will compare \method\ with multiple agents completing the task using a multi-agent plan generated by the PDDL planner. 
To enable multi-agent planning for symbolic domains, we edited PDDL domain and problem files to reflect multiple agents. 
We added an `agent' parameter to all the actions and agent-specific predicates in the domain, and initialized 2, 3, and 4 agents along with their initial states in the problem files of 20 tasks. 
% This extension requires understanding of PDDL functioning. 
Moreover, initializing another agent increases the number of possible actions at any state, as well as introducing additional state conditions, leading to a much larger search space. 
We note that this method does not support true parallelization and proposes one agent action at each timestep since it would require significant domain modifications as mentioned in section~\ref{sec: RW-MAplanning}.

3) \method: multi-agent plan with LLM-guided \helper\ and \main\ agents (Figure~\ref{fig:pipeline}). 
This method compares LLM-based division of the task into partially independent subgoals, executed in parallel, where planning for each of the $N$ agents is done as single agent PDDL problems. Our method works with single agent domains without any modifications.

\subsection{Evaluation metrics}
We measure planning time (seconds) and plan execution length (environment steps) to asses the efficacy of plans generated by different approaches described above.  
% A method that achieves lower planning time and a shorter execution length is desirable as it would be \textcolor{blue}{both} faster in finding a plan and executing it. 
The symbolic planning methods considered always result in plans that reach the goal state, so we do not compare methods by success rate. 
The proposed \method\ ensures success because the system defaults to a single agent if inferred subgoals are faulty, though planning and execution time would both increase in that case.

1) Planning Time: total time in seconds taken by the PDDL solver~\cite{fd} to explore the full state space, capped by a time limit. 
For \method, planning time is the sum of the $N-1$ \helper\ agents' PDDL planning times, \main\ PDDL planning time, and LLM inference time.

2) Execution Length: the effective length of the plan when multiple agent plans are executed in parallel. 
For single agent plans, execution length is equivalent to the best PDDL plan cost found. 
However, for multi-agent, execution length is less than or equal to that best PDDL plan cost because some sequentially planned actions can be executed in parallel by different agents.

For symbolic domains, we utilize Algorithm~\ref{alg:exec} to approximate that best possible multiagent execution length by dynamically finding a near-optimal parallel plan execution strategy. 
Algorithm~\ref{alg:exec} determines whether all $N$ agents can execute the next step in parallel by checking for conflicts. 
In case of a conflict, it executes one agent at a time, such that, the overall number of steps at the end is minimized.
Algorithm~\ref{alg:exec} finds the optimal parallel plan for 2 agents and provides a heuristic length greater than or equal to optimal for 3 or more agents.
An optimal strategy for $N$ agents requires checking for parallel execution of every possible subset of agents, a search exponential in the number of agents and actions and intractable compute time limitations.

For the embodied domain, execution length refers to the number of environment steps taken, where all agents take a step sequentially in each environment step, including navigation and interaction steps. Depending on action preconditions and execution state, we also consider some agent's action at a time step to be a `wait' or a `no operation' action.

\begin{table*}[th]
    \centering
    \setlength{\aboverulesep}{0pt}
    \caption{
        Execution length in 4 symbolic domains averaged across 4 tasks each.
        \method\ achieves comparable execution lengths as an expert human annotated subgoals (\human).
    }
    \label{tab:human}
    \begin{tabular}{kllllo}
        \bf Execution Length (steps) & \blocks & \termes & \tyre & \gripper & \textbf{Macro Avg} \\
        \method  & $15.5\pm0.4$ & $125.5\pm0.8$ & $88.3\pm1.9$ & $6.5\pm0.0$ & $58.9\pm0.8$ \\
        \human  & $16.3$ & $120.0$ & $86.8$ & $5.3$ & $57.1$
    \end{tabular}
\end{table*}

\begin{table*}[th]
    \centering
    \setlength{\aboverulesep}{0pt}
    \caption{
        Execution length and planning time in the embodied environment averaged across 4 tasks each.
        On average, \method\ achieves comparable length plans as a single agent without incurring the planning time cost of solving a large goal or a multi-agent PDDL problem.
        \method\ execution lengths are highlighted in \good{blue} when they average less than \sa\ and \bad{red} otherwise.
        Similarly, \method\ planning times are highlighted in \good{blue} when they are less than \ma\ times.
        {\footnotesize timeout} indicates reaching a time limit of 60 minutes.
    }
    \label{tab:alfworld}
    \begin{tabular}{kllllo}
        \bf Tasks & \multicolumn{5}{l}{1. Put an apple, a heated egg, and a clean fork on the dining table} \\
        & \multicolumn{5}{l}{2. Clear out the dining table by removing the plate, two bowls, and the mug} \\
        & \multicolumn{5}{l}{3. Put the apple, the egg, and the wine bottle in the fridge} \\
        & \multicolumn{5}{l}{4. Clear out the sink of the spoon, the cup, the knife, and the apple} \\
        \toprule
        \bf Execution Length (steps) & Task 1 & Task 2 & Task 3 & Task 4 & \textbf{Macro Avg} \\
        \sa  & $172.0$ & $201.0$ & $\phantom{1}99.0$ & $\phantom{1}44.0$ & $129.0$ \\
        \ma  & $189.2\pm16.7$ & $260.0\pm248.5$ & $196.7\pm88.3$ & $166.5\pm69.6$ & $203.1\pm105.8$  \\
        \method  & $\good{113.6}\pm\phantom{1}6.6$ & $\good{147.0}\pm\phantom{2}62.7$ & $\bad{204.5}\pm17.8$ & $\phantom{1}\bad{54.2}\pm14.0$ & $\bad{129.8}\pm\phantom{1}25.3$ \\
        \toprule
        \bf Planning Time (seconds) &  Task 1 & Task 2 & Task 3 & Task 4 &  \\
        \sa  & \footnotesize timeout &\footnotesize timeout & $11.3$ & \footnotesize timeout & \footnotesize 3 timeouts \\
        \ma  & \footnotesize timeout &\footnotesize timeout & \footnotesize timeout & \footnotesize timeout & \footnotesize 4 timeouts  \\
        \method  & $\good{22.0}$ & $\good{14.8}$ & $\good{18.4}$ & \footnotesize timeout & \footnotesize \good{1 timeout} \\
    \end{tabular}
\end{table*}

\subsection{Experiment Details}
PDDL planning for all approaches use \texttt{FAST-DOWNWARD} planner\footnote{https://github.com/aibasel/downward/tree/release-22.12.0} with \texttt{LAMA} alias and a search time limit of 1000 seconds for \ma\ and \sa. The \texttt{FAST-DOWNWARD} planner will keep searching until time limit for a more optimal solution.
% We found that an optimal solution was reached in this time limit in simple domains. 
For $N$ agents in \method, we provide $1000/N$ seconds as the time limit for \texttt{FAST-DOWNWARD} to solve for each agent plan.
For plan validation and extracting state condition changes with the plan execution, we use the \texttt{VAL} software.\footnote{https://github.com/KCL-Planning/VAL}
For \method's\ LLM-based modules in Figure~\ref{fig:pipeline}, we use \texttt{GPT-4o} provided as an API from OpenAI.\footnote{https://platform.openai.com/docs/models/gpt-4o\#gpt-4o}
We take the top probability response from the LLM with a temperature of 0.0 and top-$p$ of 1.0.
The results are reported for an average of 3 runs with standard deviations for \method\ because it involves stochastic LLM inference (OpenAI's API, even when temperature is 0, is nondeterministic).
We found that \texttt{FAST-DOWNWARD} does not add any variance to the pipeline, therefore we only report 1 run for all other methods, as they don't involve LLM predictions.

\section{Results}

We report results for \sa, \ma, and \method\ in symbolic and embodied domains. 
We also report a topline ablation of \method\ by replacing a \helper\ agent's LLM-based subgoal inference by a human in the symbolic domains. 
We report average execution lengths and planning times in Figure~\ref{fig:results} on each of the 5 domains with 20 problems each, for 1--4 agents. We also report the average execution cost and planning time averaged across all domains, per number of agents. Before averaging across all domains, the individual averages are normalized via min-max scaling to account for discrepancies in the magnitude of each domain's metrics. 

\textbf{\method\ takes less planning time compared to \ma\ for all domains, while coming up with more efficient multi-agent plans across all agent counts for 3/5 domains.} We include LLM generation and state editing as part of the reported planning time for \method.
We observe that \ma\ is able to find shorter plans for 2/5 domains (\blocks\ for all $N$ and \barman\ for $N=2$) compared to \method. 
\ma\ on \blocks\ is likely consistently shorter because it is a highly sequential domain, and \ma\ exhausts the state space search, thereby coming up with a shorter plan, while \method's state space search is constrained by the individual plans the LLM generates.
These 2 domains also come with defined agent-specific state conditions such as agent inventory or location, leading to a plan where multiple agents can work with different objects and not have to empty their hands before interacting with another object, or can work independently at different locations.
Improved performance in \ma\ for these domains is a byproduct of these agent-specific state conditions, not an optimization for concurrency. 
When these state conditions are missing, such as in \tyre, \ma\ is only as efficient in plan length as single agent PDDL, evidenced by the flat line for \tyre's execution length in Figure \ref{fig:results}. 
Moreover, when the domain is very complex, even with these conditions specified, \ma\ is unable to optimize the plan length within the specified time limit, as we see for \termes, where it grows worse than \sa, because adding another agent significantly increases the domain complexity (we provide the domain and examples on the website).

\textbf{Execution length and planning time for \method\ typically improves as $N$ increases.}
After normalizing execution length and planning time across domains, we find that, on average, \method\ reduces planning time by 50.1\%, 60.3\%, and 64.7\%, as well as execution length by 7.7\%, 7.1\%, and 13.2\%, for 2, 3, and 4 agents, respectively. On average, planning time decreases by 7.2\% per additional agent, showing \method's ability to break down subgoals appropriately as more agents become available.

\textbf{\method\ works with the \sa\ framework as-is, with or without agent-specific conditions specified by leveraging LLM commonsense.} 
However, \ma\ planning requires domain and problem definitions to be multi-agent compatible, and heavily depends on whether agent-specific conditions are defined.
Furthermore, temporal specification and concurrency requires additional modifications, as mentioned in Section~\ref{sec: RW-MAplanning}.

% \begin{figure}
%     \centering
%     \includegraphics[width=0.5\textwidth]{figs/human_results.pdf}
%     \caption{Human study}
%     \label{fig:human_results}
% \end{figure}

\textbf{LLM subgoal generation approximates human expert subgoal specification.}
We perform human studies to assess the quality LLM inferred subgoals in \method. 
In the first study, we ask a PDDL expert (a graduate student working with PDDL) to annotate the subgoal in PDDL goal format for 4 domains---\blocks, \termes, \tyre, \gripper---4 tasks each, sampled uniformly between tasks 1--20. 
The expert human took 70 minutes to annotate 16 tasks. 
We report execution lengths for this \human\ method in Table~\ref{tab:human}. 
\method\ performance is quite close to an expert written subgoals. 
% We also ask human unfamiliar with PDDL to annotate these tasks with natural language subgoals, however, a non-expert human found it quite challenging to write subgoals for these symbolic domains. 

% \begin{figure}
%     \centering
%     \includegraphics[width=0.5\textwidth]{figs/alfworld_results.pdf}
%     \caption{Overall execution length in AlfWorld}
%     \label{fig:res_len_alfworld}
% \end{figure}

\textbf{TwoStep advantages generalization to larger state spaces in the embodied domain.}
We report \thor\ planning time results in Table~\ref{tab:alfworld} and and execution length in Table~\ref{tab:alfworld} for the 3 approaches. 
Both \sa\ and \ma\ take a long time to plan. 
\sa\ has a large goal with several goal conditions to achieve, and \ma\ additionally has added state space complexity due to an additional agent.
In execution length, \sa\ has no variance. \ma\ and \method\ have variance in length due to random initialization of agent start location as well as random deadlock release methodology. 
The deadlocks also cause \method\ to be not achieve lower execution length than \sa\ as they add additional coordination steps. 
Moreover, \ma\ suffers a lot more with this problem, as it tries to utilize both agents back and forth, while \method\ only splits the problem with one subgoal.

\section{Conclusion}

We propose \method, a method to decompose a single agent planning problem into a multi-agent planning problem in several symbolic domains and one embodied domain. 
\method\ leverages commonsense from LLMs to effectively divide a problem between any $N$ agents for faster execution, while also preserving execution success using classical planning guarantees.
Our results show that LLM-based goal decomposition leads to faster planning time and shorter plan execution steps than the multi-agent PDDL problem on average.
We additionally show that LLM-inferred subgoals in \method\ approximate those specified by a human expert.

% \section{Limitations, Discussion and Future Work}
% We plan to apply the symbolic planning system to multi-agent AlfWorld and study the challenges in embodied environment.
% If the method succeeds with some positive rate, it may be used (along with oracle navigation) to generate embodied multi-agent planning data, and use it to train multi-agent end-to-end policy, hopefully see some emergent multi-agent behavior, such as replanning around another agent, optimize overall plan length, etc.
% This multi-agent simulation can facilitate the study of several challenges related to multi-agent planning in a realistic embodied simulation of the world, such as task allocation among multiple robots, coordination aspects including collision avoidance, action synchronization, and effective communication, adapting to dynamic environment in presence of other active agents, partial observability, and managing conflicting goals among agents while achieving collective objectives - which requires a closed-loop planning method.
% The objective would be to come up with the most effective parallel multi-agent plan.

%%%%%%%%%%%%%%%%%%%%%%%%%%%%%%%%%%%%%%%%%%%%%%%%%%%%%%%%%%%%%%%%%%%%%%%%%%%%%%%

\section*{Acknowledgements}
This work was supported by a grant from the Army Research Lab (ARL) Army AI Innovations Institute (A2I2), award number W911NF-23-2-0010.

%%%%%%%%%%%%%%%%%%%%%%%%%%%%%%%%%%%%%%%%%%%%%%%%%%%%%%%%%%%%%%%%%%%%%%%%%%%%%%%%

% References are important to the reader; therefore, each citation must be complete and correct. If at all possible, references should be commonly available publications.

\bibliographystyle{IEEEtran} % use IEEEtran.bst style
\bibliography{ref}
\newpage 
\section*{APPENDIX}
\subsection{Subgoal Generator Prompt}
\noindent \textbf{System Prompt}\\
\colorbox{blue!10}{
\begin{minipage}{0.95\columnwidth}
Main agent: agent0\\
Helper agent: agent1, ... , agent N\\
Your goal is to generate goals for agents such that they can be executed in parallel to decrease plan execution length. Generate only one clearly stated small independent subgoal for each helper agent to help the main agent complete the given task. The subgoal must be executable by a helper agent completely independently without waiting for any main agent actions to change predicates. The subgoal SHOULD NOT be interwoven with other generated subgoals or the main task, but rather run uninterrupted from inception time in PARALLEL with other subgoals.  
The subgoal should be clearly stated with unambiguous terminology. Do not use actions like assist or help, only actions CLEARLY DEFINED IN THE DOMAIN. The main goal will be augmented based on the generated subgoals, but will run in parallel with them. Do not overtake the full sequence of actions. Remember, the helper agents are only assisting the main agent and act agnostically to the main agent. If there are no subgoals that will make the task run faster, return "none". Remember that even if the domain states that there is "one" agent, you can generate as many as you'd like.
\end{minipage}
} \\ \\
\noindent \textbf{User Prompt and LLM Response}\\
Part 1: Fixed \\
% First box - Domain scenario
\colorbox{red!10}{
\begin{minipage}{0.95\columnwidth}
Example domain scenario:\\
You have 1 shaker with 3 levels, 3 shot glasses, 3 dispensers for 3 ingredients. 
The shaker and shot glasses are clean, empty, and on the table. Your left and right hands are empty. 
The first ingredient of cocktail1 is ingredient3. The second ingredient of cocktail1 is ingredient1. 
Your goal is to make 1 cocktail. 
shot1 contains cocktail1. 
\end{minipage}
}
% Second box - Agent0 steps
\colorbox{red!10}{
\begin{minipage}{0.95\columnwidth}
agent0 takes the following steps to complete the above task:\\
grasp right shot2\\
fill-shot shot2 ingredient1 right left dispenser1\\
pour-shot-to-clean-shaker shot2 ingredient1 shaker1 right l0 l1\\
clean-shot shot2 ingredient1 right left\\
fill-shot shot2 ingredient3 right left dispenser3\\
grasp left shaker1\\
pour-shot-to-used-shaker shot2 ingredient3 shaker1 right l1 l2\\
leave right shot2\\
shake cocktail1 ingredient3 ingredient1 shaker1 left right\\
pour-shaker-to-shot cocktail1 shot1 left shaker1 l2 l1
Now we have a new problem defined in this domain for which we don't have access to the single agent plan:\\You have 1 shaker with 3 levels, 3 shot glasses, 3 dispensers for 3 ingredients. 
The shaker and shot glasses are clean, empty, and on the table. Your left and right hands are empty. 
\end{minipage}
}
\colorbox{red!10}{
\begin{minipage}{0.95\columnwidth}
The first ingredient of cocktail1 is ingredient3. The second ingredient of cocktail1 is ingredient1. 
The first ingredient of cocktail2 is ingredient1. The second ingredient of cocktail2 is ingredient2. 
Your goal is to make 2 cocktails. 
shot1 contains cocktail1. shot2 contains cocktail2. 
\end{minipage}
}
% Fourth box - Agent1 subgoal
\colorbox{red!10}{
\begin{minipage}{0.95\columnwidth}
A possible agent1 subgoal looking at how the domain works based on the plan example provided for another task in this domain could be - 
agent1 subgoals: It can independently prepare cocktail1 using shot3 and shaker1. The agent will grasp shot3, fill it with ingredients, pour to shaker, shake the cocktail, and pour it to the target glass. All actions can be done while agent0 works with other containers. Therefore, agent1's clearly stated (with object names) complete and final goal condition is: contains shot1 cocktail1 and all hands are empty.
\end{minipage}
}
\colorbox{red!10}{
\begin{minipage}{0.95\columnwidth}
A possible agent2 subgoal looking at how the domain works based on the plan example provided for another task in this domain could be - 
agent2 subgoals: It can independently prepare cocktail2 using shot3 and shaker2. The agent will grasp shot3, fill it with ingredients, pour to shaker, shake the cocktail, and pour it to the target glass. All actions can be done while other agents work with other containers. Therefore, agent2's clearly stated (with object names) complete and final goal condition is: contains shot2 cocktail2 and all hands are empty.
\end{minipage}
} \\ \\
Part 2: Domain specific (\blocks) \\
\colorbox{red!10}{
\begin{minipage}{0.95\columnwidth}
Current domain scenario:\\
You have 5 blocks. 
b2 is on top of b5. 
b5 is on top of b1. 
b1 is on top of b4. 
b3 is on top of b2. 
b4 is on the table. 
b3 is clear. 
Your arm is empty. 
Your goal is to move the blocks. 
b4 should be on top of b3.
\end{minipage}
}
\colorbox{red!10}{
\begin{minipage}{0.95\columnwidth}
agent0 takes the following steps to complete the above task:\\
unstack b3 b2\\
putdown b3\\
unstack b2 b5\\
putdown b2\\
unstack b5 b1\\
putdown b5\\
unstack b1 b4\\
putdown b1\\
pickup b4\\
stack b4 b3
\end{minipage}
}
\colorbox{red!10}{
\begin{minipage}{0.95\columnwidth}
Now we have a new problem defined in this domain for which we don't have access to the signle agent plan:\\
You have 3 blocks. 
b2 is on top of b3. 
b3 is on top of b1. 
b1 is on the table. 
b2 is clear. 
Your arm is empty. 
Your goal is to move the blocks. 
b3 should be on top of b2. 
b1 should be on top of b3.
Return only one clearly stated subgoal condition for one and only one agent without explanation or steps. \\A possible subgoal looking at how the domain works based on the plan example provided for another task in this domain could be - \\ 
\end{minipage}
}
\colorbox{red!10}{
\begin{minipage}{0.95\columnwidth}
Agent 0: {previously generated subgoal would go here}\\
Agent 1: {previously generated subgoal would go here}
\\...\\
Agent N: 
\end{minipage}
}\\ \\ \\
LLM response - English Subgoal (\blocks) \\
\colorbox{green!10}{
\begin{minipage}{0.95\columnwidth}
 It can help in unstacking block b2 from b3 and putting it down on the table, while agent0 works on rearranging the other blocks. In this way, agent1 would not need to wait for agent0 and it can complete its goal independently. agent1 should also release all objects that the main agent might need for its own actions. Therefore, agent1's clearly stated (with object names) complete and final goal condition is: b2 is on the table and the arm is empty.
\end{minipage}}\\

\noindent Part 2: Domain specific (\termes) \\
\colorbox{red!10}{
\begin{minipage}{0.95\columnwidth}
Current domain scenario:\\
The robot is on a grid with 3 rows and 3 columns. \\
pos-0-0 pos-0-1 pos-0-2 \\
pos-1-0 pos-1-1 pos-1-2 \\
pos-2-0 pos-2-1 pos-2-2 \\
The robot is at pos-2-0. \\
The depot for new blocks is at pos-2-0. 
The maximum height of blocks is 2. 
Your goal is to build blocks so that the height at pos-1-1 is 2. 
You cannot have an unplaced block at the end.
\end{minipage}
}
\colorbox{red!10}{
\begin{minipage}{0.95\columnwidth}
agent0 takes the following steps to complete the above task:\\
create-block pos-2-0\\
place-block pos-2-0 pos-1-0 n0 n1\\
create-block pos-2-0\\
move pos-2-0 pos-2-1 n0\\
place-block pos-2-1 pos-1-1 n0 n1\\
move pos-2-1 pos-2-0 n0\\
create-block pos-2-0\\
move-up pos-2-0 n0 pos-1-0 n1\\
place-block pos-1-0 pos-1-1 n1 n2\\
move-down pos-1-0 n1 pos-2-0 n0\\
remove-block pos-2-0 pos-1-0 n1 n0\\
destroy-block pos-2-0
\end{minipage}
}
\colorbox{red!10}{
\begin{minipage}{0.95\columnwidth}
Now we have a new problem defined in this domain for which we don't have access to the signle agent plan:\\
The robot is on a grid with 4 rows and 3 columns. \\
pos-0-0 pos-0-1 pos-0-2 \\
pos-1-0 pos-1-1 pos-1-2 \\
pos-2-0 pos-2-1 pos-2-2 \\
pos-3-0 pos-3-1 pos-3-2 \\
The robot is at pos-2-0. \\
The depot for new blocks is at pos-2-0. 
The maximum height of blocks is 3. 
Your goal is to build blocks so that the height at pos-1-2 is 3. 
You cannot have an unplaced block at the end.
Return only one clearly stated subgoal condition for one and only one agent without explanation or steps. \\
\end{minipage}
}
\colorbox{red!10}{
\begin{minipage}{0.95\columnwidth}
A possible subgoal looking at how the domain works based on the plan example provided for another task in this domain could be - \\ 
Agent 0: {previously generated subgoal would go here}\\
Agent 1: {previously generated subgoal would go here}
\\...\\
Agent N: 
\end{minipage}
}
\\ \\ \\
LLM response - English Subgoal (\termes) \\
\colorbox{green!10}{
\begin{minipage}{0.95\columnwidth}
It can help in creating a block at the depot pos-2-0, then move it to pos-1-2, while agent0 prepares other blocks using other positions. In this way, agent1 would not need to wait for agent0 and it can complete its goal independently. agent1 should also ensure that there are no unplaced blocks at the end. Therefore, agent1's clearly stated (with object names) complete and final goal condition is: pos-1-2 has a block of height n1 and all unplaced blocks are destroyed.
\end{minipage}}\\ \\
\noindent Part 2: Domain specific (\tyre) \\
\colorbox{red!10}{
\begin{minipage}{0.95\columnwidth}
Current domain scenario:\\
You have a jack, a pump, a wrench, a boot, 2 hubs, 2 nuts, 2 flat tyres, and 2 intact tyres. 
The jack, pump, wrench, and intact tyres are in the boot. 
The boot is unlocked but is closed. 
The intact tyres are not inflated. 
The flat tyres are on the hubs. 
The hubs are on the ground. 
The nuts are tight on the hubs. 
The hubs are fastened. 
Your goal is to replace flat tyres with intact tyres on the hubs. Intact tyres should be inflated. The nuts should be tight on the hubs. The flat tyres, wrench, jack, and pump should be in the boot. The boot should be closed.
\end{minipage}
}
\colorbox{red!10}{
\begin{minipage}{0.95\columnwidth}
agent0 takes the following steps to complete the above task:\\
loosen nuts2 the-hub2 \\
jack-up the-hub2 \\
undo nuts2 the-hub2 \\
remove-wheel w2 the-hub2 \\
loosen nuts1 the-hub1 \\
jack-up the-hub1 \\
undo nuts1 the-hub1 \\
remove-wheel w1 the-hub1 \\
inflate r2 \\
inflate r1 \\
open boot \\
fetch r1 boot \\
put-on-wheel r1 the-hub1 \\
do-up nuts1 the-hub1\\
jack-down the-hub1\\
tighten nuts1 the-hub1\\
fetch r2 boot\\
put-on-wheel r2 the-hub2\\
do-up nuts2 the-hub2\\
jack-down the-hub2\\
tighten nuts2 the-hub2\\
put-away w1 boot\\
put-away w2 boot\\
close boot
\end{minipage}
}
\colorbox{red!10}{
\begin{minipage}{0.95\columnwidth}
Now we have a new problem defined in this domain for which we don't have access to the signle agent plan:\\
You have a jack, a pump, a wrench, a boot, 1 hubs, 1 nuts, 1 flat tyres, and 1 intact tyres. 
The jack, pump, wrench, and intact tyres are in the boot. 
The boot is unlocked but is closed. 
The intact tyres are not inflated. 
The flat tyres are on the hubs. 
The hubs are on the ground. 
The nuts are tight on the hubs. 
The hubs are fastened. 
Your goal is to replace flat tyres with intact tyres on the hubs. Intact tyres should be inflated. The nuts should be tight on the hubs. The flat tyres, wrench, jack, and pump should be in the boot. The boot should be closed.
\end{minipage}
}
\colorbox{red!10}{
\begin{minipage}{0.95\columnwidth}
A possible agent1 subgoal looking at how the domain works based \\ \\
Return only one clearly stated subgoal condition for one and only one agent without explanation or steps. \\A possible subgoal looking at how the domain works based on the plan example provided for another task in this domain could be - \\ 
Agent 0: {previously generated subgoal would go here}\\
Agent 1: {previously generated subgoal would go here}
\\...\\
Agent N: 
\end{minipage}
}\\ \\
LLM response - English Subgoal (\tyre) \\
\colorbox{green!10}{
\begin{minipage}{0.95\columnwidth}
 It can help in inflating the intact tyre while agent0 works on removing the flat tyre from the hub. In this way, agent1 would not need to wait for agent0 and it can complete its goal independently. agent1 should also release all objects that the main agent might need for its own actions. Therefore, agent1's clearly stated (with object names) complete and final goal condition is: the intact tyre is inflated and all hands are empty.
\end{minipage}}\\ \\ \\

\noindent Part 2: Domain specific (\gripper) \\
\colorbox{red!10}{
\begin{minipage}{0.95\columnwidth}
Current domain scenario:\\
You control 1 robot, each robot has a left gripper and a right gripper. 
There are 4 rooms and 6 balls. 
robot1 is in room3.
ball1 is in room3. ball2 is in room1. ball3 is in room3. ball4 is in room2. ball5 is in room4. ball6 is in room4. 
The robots' grippers are free. 
Your goal is to transport the balls to their destinations. 
ball1 should be in room4. 
ball2 should be in room1. 
ball3 should be in room1. 
ball4 should be in room2. 
ball5 should be in room1. 
ball6 should be in room1.
agent0 takes the following steps to complete the above task:\\
pick robot1 ball1 room3 lgripper2\\
move robot1 room3 room1\\
move robot1 room1 room4\\
pick robot1 ball5 room4 rgripper2\\
drop robot1 ball1 room4 lgripper2\\
pick robot1 ball6 room4 lgripper2\\
move robot1 room4 room1\\
drop robot1 ball6 room1 lgripper2\\
\end{minipage}
}
\colorbox{red!10}{
\begin{minipage}{0.95\columnwidth}
drop robot1 ball5 room1 rgripper2\\
move robot1 room1 room3\\
pick robot1 ball3 room3 lgripper2\\
move robot1 room3 room1\\
drop robot1 ball3 room1 lgripper2
\end{minipage}
}
\colorbox{red!10}{
\begin{minipage}{0.95\columnwidth}
Now we have a new problem defined in this domain for which we don't have access to the signle agent plan:\\
You control 1 robots, each robot has a left gripper and a right gripper. 
There are 2 rooms and 2 balls. 
robot1 is in room1.
ball2 is in room1. ball1 is in room1. 
The robots' grippers are free. 
Your goal is to transport the balls to their destinations. 
ball1 should be in room2. 
ball2 should be in room2.
\end{minipage}
}
\colorbox{red!10}{
\begin{minipage}{0.95\columnwidth}
Return only one clearly stated subgoal condition for one and only one agent without explanation or steps. \\A possible subgoal looking at how the domain works based on the plan example provided for another task in this domain could be - \\ 
Agent 0: {previously generated subgoal would go here}\\
Agent 1: {previously generated subgoal would go here}
\\...\\
Agent N: 
\end{minipage}
}\\ \\
LLM response - English Subgoal (\gripper) \\
\colorbox{green!10}{
\begin{minipage}{0.95\columnwidth}
 It can help in picking up ball1 in room1 and transporting it to room2. In this way, agent1 would not need to wait for agent0 and it can complete its goal independently. agent1 should also release all objects that the main agent might need for its own actions. Therefore, agent1's clearly stated (with object names) complete and final goal condition is: ball1 is in room2 and all grippers are free.
\end{minipage}}\\ \\

\subsection{Subgoal Translator Prompt}

\noindent \textbf{System Prompt}\\
\colorbox{blue!10}{\begin{minipage}{0.95\columnwidth}
I want you to solve planning problems. Provide me with the PDDL goal that describes the new planning goal directly without further explanations. Make sure to provide only non-conflicting, necessary, and final goal conditions mentioned in the given goal.
\end{minipage}} \\ \\
\noindent \textbf{User Prompt and LLM Response}\\
Domain and Problem specific (\blocks) \\
\colorbox{red!10}{
\begin{minipage}{0.95\columnwidth}
The PDDL problem and its initial conditions are given as: \\
(define (problem BW-rand-3)\\
(:domain blocksworld-4ops)\\
(:objects b1 b2 b3 )\\
(:init\\
(arm-empty)\\
(on-table b1)\\
(on b2 b3)\\
(on b3 b1)\\
(clear b2)\\
An example planning goal for this problem:  \\
Your goal is to move the blocks. 
)
\end{minipage}
}
\colorbox{red!10}{
\begin{minipage}{0.95\columnwidth}
b3 should be on top of b2. 
b1 should be on top of b3. 
\end{minipage}
}
\colorbox{red!10}{
\begin{minipage}{0.95\columnwidth}
The PDDL goal for the example planning goal:  \\
(:goal\\
(and\\
(on b3 b2)\\
(on b1 b3))\\
)
\end{minipage}
}
\colorbox{red!10}{
\begin{minipage}{0.95\columnwidth}
New planning goal for the same problem:\\
Your goal is: b2 is on the table and the arm is empty. 
\end{minipage}
}
\colorbox{red!10}{
\begin{minipage}{0.95\columnwidth}
The PDDL goal for the new planning goal:
\end{minipage}
} \\ \\
LLM response - PDDL Subgoal (\blocks) \\
\colorbox{green!10}{
\begin{minipage}{0.95\columnwidth}
(:goal \\
(and \\
(on-table b2) \\
(arm-empty)) \\
)
\end{minipage}}\\ \\
\noindent Domain and Problem specific (\termes) \\
\colorbox{red!10}{
\begin{minipage}{0.95\columnwidth}
The PDDL problem and its initial conditions are given as:  \\
(define (problem termes-00038-0036-4x3x3-random\_towers\_4x3\_3\_1\_3) \\
(:domain termes) \\
; termes-00038-0036-4x3x3-random\_towers\_4x3\_3\_1\_3 \\
; Initial state: \\
;  0   0  R0D  0 \\
;  0   0   0   0 \\
;  0   0   0   0 \\
; Goal state: \\
;  0   0   0   0 \\
;  0   0   0   0 \\
;  0   3   0   0 \\
; Maximal height: 3
\end{minipage}
}
\colorbox{red!10}{
\begin{minipage}{0.95\columnwidth}
(:objects \\
    n0 - numb \\
    n1 - numb \\
    n2 - numb \\
    n3 - numb \\
    pos-0-0 - position \\
    pos-0-1 - position \\
.... \\
    pos-3-0 - position \\
    pos-3-1 - position \\
    pos-3-2 - position \\
    (:init \\
    (height pos-0-0 n0) \\
..... \\
    (height pos-3-0 n0) \\
    (height pos-3-1 n0) \\
    (height pos-3-2 n0) \\
    (at pos-2-0) \\
    (SUCC n1 n0) \\
    (SUCC n2 n1) \\
)
\end{minipage}
}
\colorbox{red!10}{
\begin{minipage}{0.95\columnwidth}
    (SUCC n3 n2) \\
    (NEIGHBOR pos-0-0 pos-1-0) \\
    (NEIGHBOR pos-0-0 pos-0-1) \\
    (NEIGHBOR pos-0-1 pos-1-1) \\
    (NEIGHBOR pos-0-1 pos-0-0) \\
    (NEIGHBOR pos-0-1 pos-0-2) \\
    (NEIGHBOR pos-0-2 pos-1-2) \\
..... \\
    (NEIGHBOR pos-3-2 pos-2-2) \\
    (NEIGHBOR pos-3-2 pos-3-1) \\
    (IS-DEPOT pos-2-0) \\
)
\end{minipage}
}
\colorbox{red!10}{
\begin{minipage}{0.95\columnwidth}
An example planning goal for this problem:   \\
Your goal is to build blocks so that the height at pos-1-2 is 3. 
You cannot have an unplaced block at the end.
The PDDL goal for the example planning goal:   \\
(:goal \\
(and \\
    (height pos-0-0 n0) \\
    (height pos-0-1 n0) \\
    (height pos-0-2 n0) \\
    (height pos-1-0 n0) \\
    (height pos-1-1 n0) \\
    (height pos-1-2 n3) \\
    (height pos-2-0 n0) \\
    (height pos-2-1 n0) \\
    (height pos-2-2 n0) \\
    (height pos-3-0 n0) \\
    (height pos-3-1 n0) \\
    (height pos-3-2 n0) \\
    (not (has-block)) \\
\end{minipage}
}
\colorbox{red!10}{
\begin{minipage}{0.95\columnwidth}
New planning goal for the same problem: \\
Your goal is: pos-1-2 has a block of height n1 and the robot is at pos-2-0.
\end{minipage}
}
\colorbox{red!10}{
\begin{minipage}{0.95\columnwidth}
The PDDL goal for the new planning goal:
\end{minipage}
} \\ \\
LLM response - PDDL Subgoal (\termes) \\
\colorbox{green!10}{
\begin{minipage}{0.95\columnwidth}
(:goal \\
(and \\
    (height pos-0-0 n0) \\
    (height pos-0-1 n0) \\
    (height pos-0-2 n0) \\
    (height pos-1-0 n0) \\
    (height pos-1-1 n0) \\
    (height pos-1-2 n1) \\
    (height pos-2-0 n0) \\
    (height pos-2-1 n0) \\
    (height pos-2-2 n0) \\
    (height pos-3-0 n0) \\
    (height pos-3-1 n0) \\
    (height pos-3-2 n0) \\
    (not (has-block)) \\
)
)
\end{minipage}}\\ \\
\newpage
\noindent
Domain and Problem specific (\tyre) \\
\colorbox{red!10}{
\begin{minipage}{0.95\columnwidth}
The PDDL problem and its initial conditions are given as: \\
(define (problem tyreworld-1)\\
(:domain tyreworld)\\
(:objects \\
wrench jack pump - tool\\
the-hub1 - hub\\
nuts1 - nut\\
boot - container\\
r1 w1 - wheel\\
)\\
(:init\\
(in jack boot)\\
(in pump boot)\\
(in wrench boot)\\
(unlocked boot)\\
(closed boot)\\
(intact r1)\\
(in r1 boot)\\
(not-inflated r1)\\
(on w1 the-hub1)\\
(on-ground the-hub1)\\
(tight nuts1 the-hub1)\\
(fastened the-hub1)\\
) 
\\ \\
An example planning goal for this problem:  \\
Your goal is to replace flat tyres with intact tyres on the hubs. Intact tyres should be inflated. The nuts should be tight on the hubs. The flat tyres, wrench, jack, and pump should be in the boot. The boot should be closed. 
\\ \\
The PDDL goal for the example planning goal:  \\
(:goal\\
(and\\
(on r1 the-hub1)\\
(inflated r1)\\
(tight nuts1 the-hub1)\\
(in w1 boot)\\
(in wrench boot)\\
(in jack boot)\\
(in pump boot)\\
(closed boot)\\
)
) 
\\ \\
New planning goal for the same problem:\\
 Your goal is: the intact tyre is inflated and the pump is in the boot. 
 \\ \\
The PDDL goal for the new planning goal:
\end{minipage}}\\ \\ \\
\newpage \noindent
LLM response - PDDL Subgoal (\tyre) \\
\colorbox{green!10}{
\begin{minipage}{0.95\columnwidth}
(:goal \\
(and \\
(inflated r1) \\
)
\end{minipage}}\\ \\ \\
\noindent
Domain and Problem specific (\gripper) \\
\colorbox{red!10}{
\begin{minipage}{0.95\columnwidth}
The PDDL problem and its initial conditions are given as: \\
(define (problem gripper-2-2-2)\\
(:domain gripper-strips)\\
(:objects robot1 - robot\\
rgripper1 lgripper1 - gripper\\
room1 room2 - room\\
ball1 ball2 - object)\\
(:init\\
(at-robby robot1 room1)\\
(free robot1 rgripper1)\\
(free robot1 lgripper1)\\
(at ball1 room1)\\
(at ball2 room1)\\
) 
\\ \\
An example planning goal for this problem:  \\
Your goal is to transport the balls to their destinations. 
ball1 should be in room2. 
ball2 should be in room2. 
\\ \\
The PDDL goal for the example planning goal:  \\
(:goal\\
(and\\
(at ball1 room2)\\
(at ball2 room2)\\
)
) 
\\ \\
New planning goal for the same problem:\\
 Your goal is: ball1 is in room2 and all grippers are free. 
\\ \\
The PDDL goal for the new planning goal:
\end{minipage}}\\ \\ \\
LLM response - PDDL Subgoal (\gripper) \\
\colorbox{green!10}{
\begin{minipage}{0.95\columnwidth}
(:goal \\
(and \\
(at ball1 room2)\\
(free robot1 rgripper1)\\
(free robot1 lgripper1)\\
)
\end{minipage}}
\newpage
Domain and Problem specific (\barman) \\
\colorbox{red!10}{
\begin{minipage}{0.95\columnwidth}
The PDDL problem and its initial conditions are given as: \\
(define (problem prob)\\
 (:domain barman)\\
 (:objects \\
      shaker1 - shaker\\
      left right - hand\\
      shot1 shot2 shot3 shot4 - shot\\
      ingredient1 ingredient2 ingredient3 - ingredient\\
      cocktail1 cocktail2 cocktail3 - cocktail\\
      dispenser1 dispenser2 dispenser3 - dispenser\\
      l0 l1 l2 - level\\
)
\end{minipage}
}
\colorbox{red!10}{
\begin{minipage}{0.95\columnwidth}
 (:init \\
  (ontable shaker1)\\
  (ontable shot1)\\
  (ontable shot2)\\
  (ontable shot3)\\
  (ontable shot4)\\
  (dispenses dispenser1 ingredient1)\\
  (dispenses dispenser2 ingredient2)\\
  (dispenses dispenser3 ingredient3)\\
  (clean shaker1)\\
  (clean shot1)\\
  (clean shot2)\\
  (clean shot3)\\
  (clean shot4)\\
  (empty shaker1)\\
  (empty shot1)\\
  (empty shot2)\\
  (empty shot3)\\
  (empty shot4)\\
  (handempty left)\\
  (handempty right)\\
  (shaker-empty-level shaker1 l0)\\
  (shaker-level shaker1 l0)\\
  (next l0 l1)\\
  (next l1 l2)\\
  (cocktail-part1 cocktail1 ingredient1)\\
  (cocktail-part2 cocktail1 ingredient3)\\
  (cocktail-part1 cocktail2 ingredient2)\\
  (cocktail-part2 cocktail2 ingredient3)\\
  (cocktail-part1 cocktail3 ingredient1)\\
  (cocktail-part2 cocktail3 ingredient2)\\
) 
\end{minipage}
}
\colorbox{red!10}{
\begin{minipage}{0.95\columnwidth}
An example planning goal for this problem:  \\
Your goal is to make 3 cocktails. 
shot1 contains cocktail1. shot2 contains cocktail3. shot3 contains cocktail2. 
\end{minipage}
}
\colorbox{red!10}{
\begin{minipage}{0.95\columnwidth}
The PDDL goal for the example planning goal:  \\
(\\
(and\\
      (contains shot1 cocktail1)\\
      (contains shot2 cocktail3)\\
      (contains shot3 cocktail2)\\
)) 
\end{minipage}
}
\colorbox{red!10}{
\begin{minipage}{0.95\columnwidth}
New planning goal for the same problem:\\
 Your goal is: contains shot1 cocktail1 and all hands are empty. 
\end{minipage}
}
\colorbox{red!10}{
\begin{minipage}{0.95\columnwidth}
The PDDL goal for the new planning goal:
\end{minipage}
} \\ \\
LLM response - PDDL Subgoal (\barman) \\
\colorbox{green!10}{
\begin{minipage}{0.95\columnwidth}
 (:goal\\
 (and\\
 (contains shot1 cocktail1)\\
 (handempty left)\\
 (handempty right)\\
 ) )
\end{minipage}}\\ \\

\end{document}